\documentclass[conference]{IEEEtran}
\usepackage{graphicx} 
\usepackage[font=scriptsize]{caption}
\usepackage{subcaption}
\usepackage{amsmath}
\usepackage{authblk}
\usepackage{hyperref}
\usepackage{blindtext}
\usepackage[english]{babel}
\usepackage[utf8]{inputenc}
\usepackage[dvipsnames]{xcolor}
\usepackage[]{algorithm2e}
\usepackage{algpseudocode}
\usepackage{listings}
\usepackage{cancel}
\usepackage{ulem}
\usepackage{float}
\usepackage{url}
\usepackage{breakurl}

\author{
   Assylzhan Izbassar\\
  School of Information Technology and Engineering\protect\\Kazakh-British Technical University\\Almaty, Kazakhstan\\
  \texttt{a.izbassar@kbtu.kz}
  \and
  Pakizar Shamoi\\
  School of Information Technology and Engineering\protect\\Kazakh-British Technical University\\Almaty, Kazakhstan\\
  \texttt{p.shamoi@kbtu.kz}
}
\title{Image-Based Dietary Assessment: A Healthy Eating Plate Estimation System}

\begin{document}



\maketitle
\IEEEpeerreviewmaketitle

\begin{abstract}
The nutritional quality of diets has significantly deteriorated over the past two to three decades, a decline often underestimated by the people. This deterioration, coupled with a hectic lifestyle, has contributed to escalating health concerns. Recognizing this issue, researchers at Harvard have advocated for a balanced nutritional plate model to promote health. Inspired by this research, our paper introduces an innovative Image-Based Dietary Assessment system aimed at evaluating the healthiness of meals through image analysis. Our system employs advanced image segmentation and classification techniques to analyze food items on a plate, assess their proportions, and calculate the meal's adherence to Harvard's healthy eating recommendations. This approach leverages machine learning and nutritional science to empower individuals with actionable insights for healthier eating choices. Our four-step framework involves segmenting the image, classifying the items, conducting a nutritional assessment based on the Harvard Healthy Eating Plate research, and offering tailored recommendations. The prototype system has shown promising results in promoting healthier eating habits by providing an accessible, evidence-based tool for dietary assessment. 

 \end{abstract}


\section{Introduction}

The nutrition of modern humans is very different from what they were eating several decades ago. We face significant challenges in maintaining healthy eating habits, resulting in inadequate nutrition, diet, and associated health issues. According to recent research of the National Health and Nutrition Examination Survey (NHANES) conducted by the Centers for Disease Control and Prevention (CDC), a significant portion of the American population does not meet the recommended intake for dietary fiber \cite{nhanes}. The Dietary Guidelines for Americans recommend 25 grams and 38 grams of fiber daily for women and men, respectively. Based on the NHANES data, it has been reported that only a small percentage of Americans meet these recommendations. For instance, the studies of \cite{demmer} showed that in every group of 2379 girls aged 9-13, 14-18, and 19-20, and in each of those ages, they found that more than 90\% did not consume the needed amount of fruits, vegetables or dairy. Hence, we can safely assume that people do not pay enough attention to what they eat daily. Furthermore, the modern food environment, characterized by easy access to energy-dense, nutrient-poor foods, contributes to overeating and inadequate nutrient intake \cite{hall}. Then, studies by \cite{poti} and \cite{rehm} highlight the prevalence of excessive consumption of added sugars, unhealthy fats, and processed foods, leading to imbalanced diets and increased risk of chronic diseases.

Harvard's Research on the "Healthy Eating Plate" offers evidence-based guidelines to promote healthier eating habits \cite{harvard}. According to their research, the "Healthy Eating Plate" emphasizes consuming whole grains, fruits, vegetables, lean proteins, and healthy fats. This approach aligns with recommendations from renowned institutions, such as the World Health Organization and the Dietary Guidelines for Americans, providing a comprehensive solution to address inadequate eating. Studies have explored the impact of the Healthy Eating Plate on dietary behaviors. Research by \cite{mekonnen} found that individuals who followed the Healthy Eating Plate guidelines had higher intakes of fruits, vegetables, and whole grains and lower intakes of unhealthy fats and added sugars. Moreover, a study by \cite{sacks} demonstrated that adherence to a healthy eating pattern, similar to the "Healthy Eating Plate", was associated with a reduced risk of major chronic diseases.

In this particular context, another issue arises as individuals, despite knowing about optimal nutrition, encounter a substantial hurdle in incorporating it into their dietary practices. The transition from consuming fast foods to a healthful diet presents itself as a challenge \cite{popkin}. Moreover, calculating grams and calories for each food item becomes imperative to achieve dietary equilibrium. While nutritionists can offer guidance with these computations, automating this process would yield immense benefits. Thus, the advent of a system capable of evaluating the balance of a plate merely through image analysis would prove invaluable. This article delves into the exploration of object recognition in images and its practical implementation for assessing plate balance, drawing insights from the guidelines provided by the Harvard Research Center.


The current paper aims to use image processing for dietary assessment based on Harvard’s nutritional recommendations for a healthy plate. The contributions of this study are integrating official Harvard Medical School Recommendations with ML and Nutrition coordination - providing recommendations to improve the plate health level.

The paper is organized as follows. Section I is this Introduction. A review of earlier related studies is presented in Section II. The several techniques we employ in this work, such as image segmentation and classification and explanations of data collection, are all included in Section III. The sample application and experimental findings are presented in Section IV. Finally, concluding remarks are given in Section V, which suggests future improvements.

\section{Related Work}
\subsection{Image processing for food detection and dietary assessment}

This section provides an overview of research related to plate health level estimation. Several related works have explored the use of various algorithms and techniques in food image analysis. Phanich's research focused on developing a food recommendation system for diabetic patients using clustering analysis, specifically utilizing the k-means algorithm \cite{phanich}. The other studies utilized the k-means algorithm focusing on food detection based on the HSV color model \cite{pham} and the Sobel operator and k-means clustering for automatic image segmentation in volume measurement systems for food products \cite{siswantoro}.  The other study used Faster R-CNN and YOLO for object detection and measuring calorie and nutrition content from food images \cite{shimoda}. These research works collectively demonstrate the significance of k-means clustering and neural networks in analyzing food images and their application in determining the healthy level of food on plates through image-based approaches. Some research about plate food recognition uses different approaches using phone-taken images. For instance, \cite{kawano_real-time_2013} recognizes foods in the plate by using linear SVM and fast \(\chi^2\) kernel.

\subsection{Nutritional recommendations: Overview}

The Harvard Study on healthy eating of modern humans is a universal recommendation for a balanced diet. This approach helps to maintain optimal bodily functions, improve overall well-being, and effectively regulate cholesterol levels.

\begin{figure}[htp]
    \centering
    \includegraphics[width=7cm]{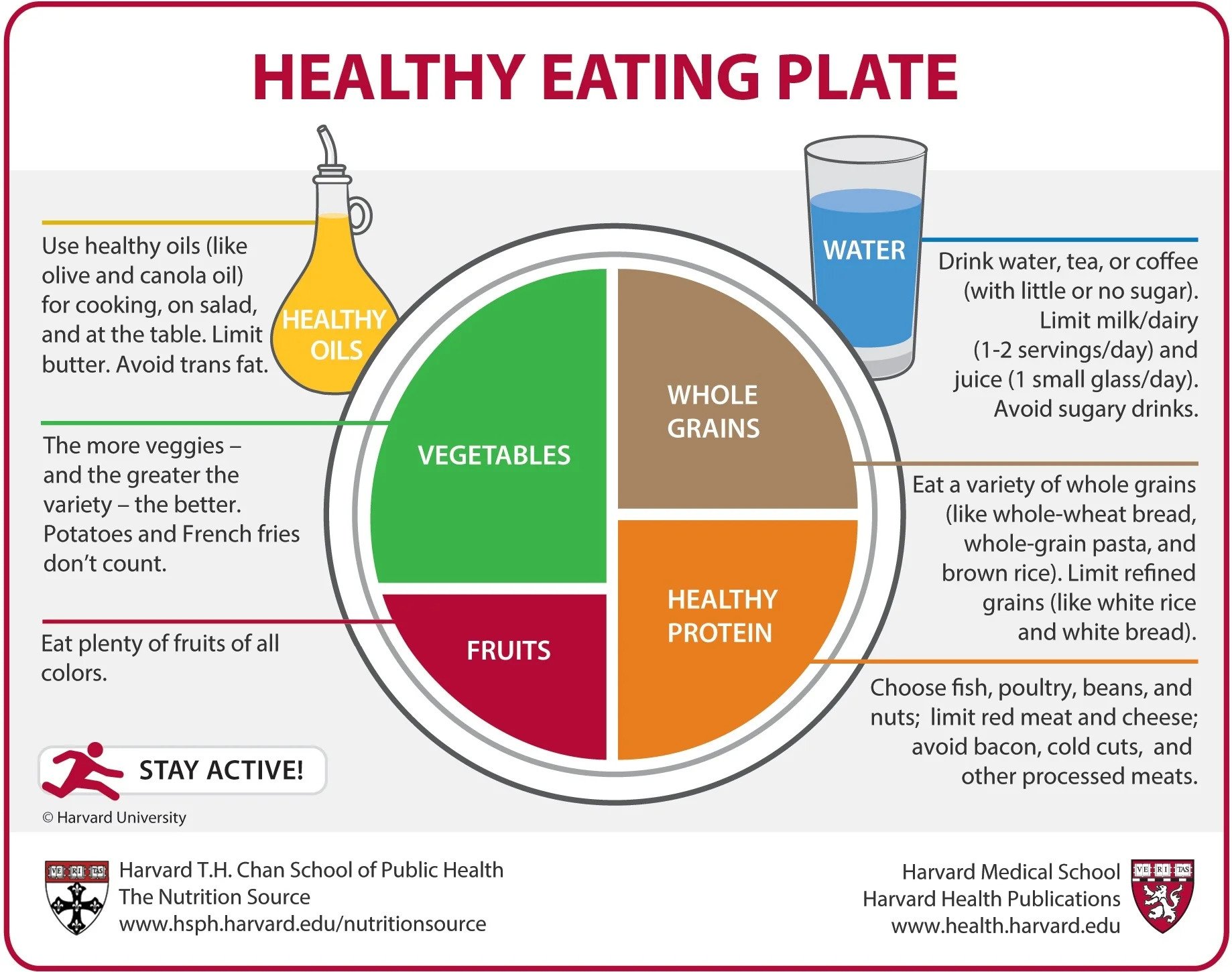}
    \caption{Healthy plate recommendation. For more information, please see The Nutrition Source, Department of Nutrition, Harvard T.H. Chan School of Public Health, www.thenutritionsource.org, and Harvard Health Publications, www.health.harvard.edu.}
    \label{fig:health_plate}
\end{figure}

In Fig. \ref{fig:health_plate}, Harvard experts share seven key tips for healthy eating. They suggest filling half our plates with colorful fruits and veggies \cite{harvard_veg_fruit}, which can help lower blood pressure, reduce heart disease risk, prevent some cancers, regulate cholesterol, and more. They also advise choosing various colors for a broader range of vitamins and minerals. They recommend excluding potatoes from the vegetable category because they can affect blood sugar levels.


\begin{figure*}
 \includegraphics[width=\textwidth]{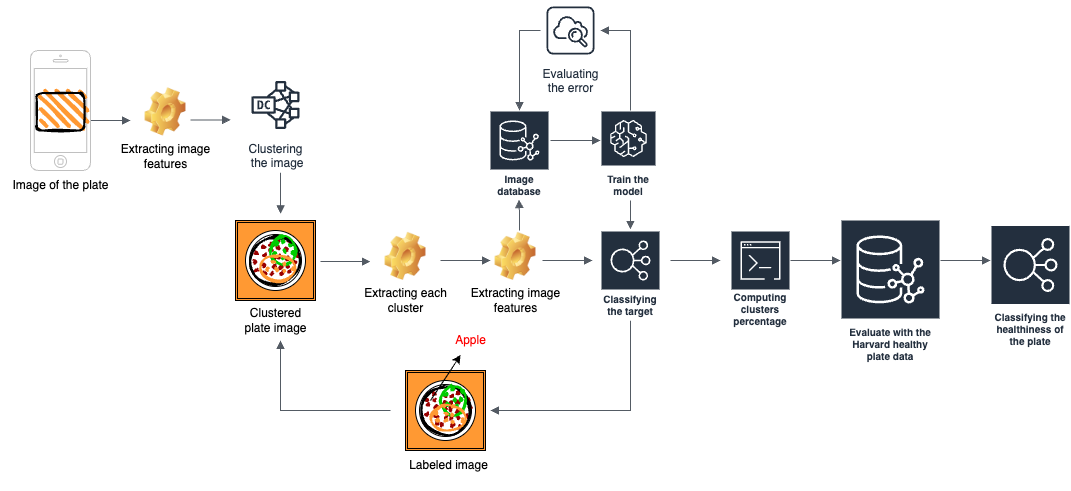}
    \caption{Clustering of the plate}
    \label{fig:methodology_figure}
\end{figure*}

Another important part of a balanced plate is protein, making up a quarter of it. Protein is crucial for many body parts, like muscles, skin, and bones. Adults should aim for about 0.8 grams of protein per kilogram of body weight each day \cite{protein}, which for a 70-kilogram person is around 56 grams or a quarter of the daily plate. Great protein sources include fish, lean meats, beans, lentils, and nuts. However, too much red meat can raise the risk of heart problems. So, it's good to mix healthy options like beans, nuts, or fish to keep things balanced and reduce those risks \cite{cardiovascular}.


Moreover, whole grains like wheat, buckwheat, and rice in our meals are advised for their rich nutrients \cite{grains}, including vitamins B and E, iron, copper, and zinc. Their fiber content aids in stabilizing blood sugar levels, reducing cholesterol, and promoting smooth digestion, enhancing our overall health.


In addition to these fundamental components of a nourishing plate, it is essential to incorporate healthy vegetable oils into our diet \cite{fats}. Optimal choices include olive oil, canola oil, soybean oil, corn oil, sunflower oil, and peanut oil. These oils offer valuable health benefits and contribute to a balanced eating plan. Conversely, it is important to avoid consuming partially hydrogenated oils, as they contain harmful trans fats that can negatively impact our well-being. By opting for healthier vegetable oils, we can enhance the nutritional value of our meals while promoting overall wellness.


\section{Methods}
\label{sec:methodology}
In this paper, we propose an image understanding system that takes an image of the food plate as an input and predicts its health level. We have a four-step framework: segment image, classify items, perform a nutritional assessment based on Harvard Healthy Plate research, and provide recommendations. 

As shown in Figure \ref{fig:methodology_figure}, the camera image is initially processed into a 2D format for clustering algorithms like mean-shifting and segmentation. After obtaining masks for food items, each object's class is determined through feature extraction. Harvard studies are referenced to estimate combined fractions, followed by assessing the plate's overall performance and guidance for improvement.


\subsection{Experimental settings}

\begin{enumerate}
    \item Placement Guidelines for Objects on a Plate:
    \begin{itemize}
        \item For an experiment, we selected a pristine white plate with a radius of 10 cm;
        \item The maximum number of objects allowed on the plate should range between 7 to 8 pieces;
        \item Objects should not overlap each other, such as placing a fried egg on top of a cutlet. Our current model does not account for such arrangements;
        \item Small grain additives must be provided separately;
        \item Objects should not be filled with any sauces.
    \end{itemize}
    
    \item Object Requirements for Plate Composition:
    \begin{itemize}
        \item Only food items are permitted on the plate. 
        \item The current model does not consider complex foods like cakes. 
        \item Objects should not be shredded; it's preferable to maintain their original form. For instance, an apple should not be cut into fine julienne strips, and peas, typically measured in grams, should not be sliced;
        \item Solid objects should remain as individual pieces, while items measured in grams should be represented as such. For example, grains should not be shown individually, whereas a fish fillet can be depicted as a single piece;
        \item Objects can be shown in a fried state.
    \end{itemize}    
    \item Image Requirements:
    \begin{itemize}
        \item Photos must be captured from a top-down perspective.
    \end{itemize}
\end{enumerate}

\subsection{Data Collection}
\subsubsection{Dataset of bowls}
We sourced our primary dataset from the Roboflow platform \cite{plate_bowl_spyne}. This dataset consisted of 279 images featuring plates and bowls. For our experiment, we selected approximately 15 images from this dataset. Additionally, we supplemented our dataset with images of other healthy plate items obtained from freely accessible sources.

\subsubsection{Train data of a classification algorithm}

\begin{figure}[bt]
\centering
\includegraphics[width=.45\textwidth]{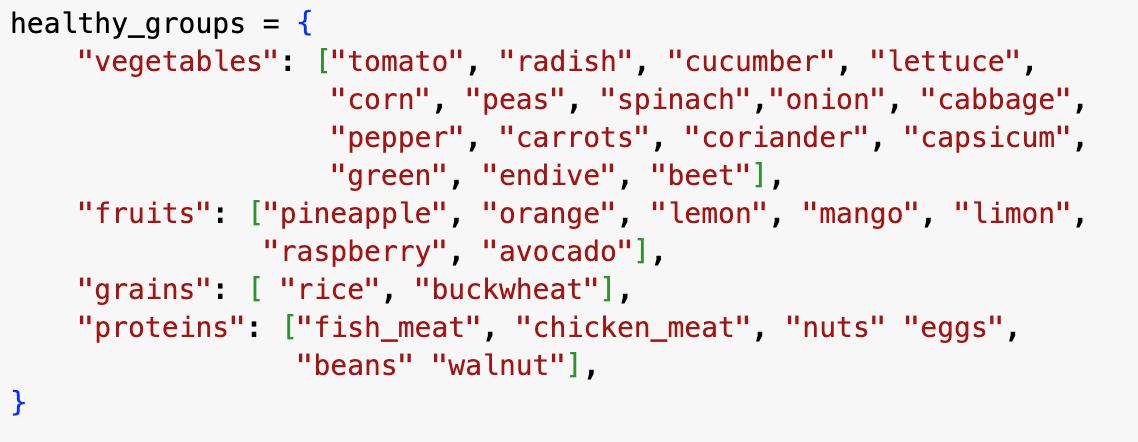}
\caption{Categories of healthy objects}
\label{fig:categories}
\end{figure}

The main part of the working algorithm is to prepare the proper dataset. We have categorized the training image database into four main groups: fruits, vegetables, proteins, and whole grains. Each category comprises specific types of fruits, vegetables, proteins, and whole grains, as shown in Figure \ref{fig:categories}.

\subsection{Image Processing Techniques}
To analyze food plate images, we employ image segmentation and classification techniques. The eating plate images were originally in various sizes. The image size has been normalized to (256, 256), and the image mode has been set to RGB.

\subsubsection{Image Segmentation}

It involves dividing an image into meaningful sections based on color, texture, or other visual features. This helps distinguish objects from the background, aiding in analysis. Adapted from \cite{anthimopoulos}, our approach includes color conversion, clustering, region merging, and background subtraction for plate food detection.

\begin{figure}[tb]
    \centering
    \includegraphics[width=.45\textwidth]{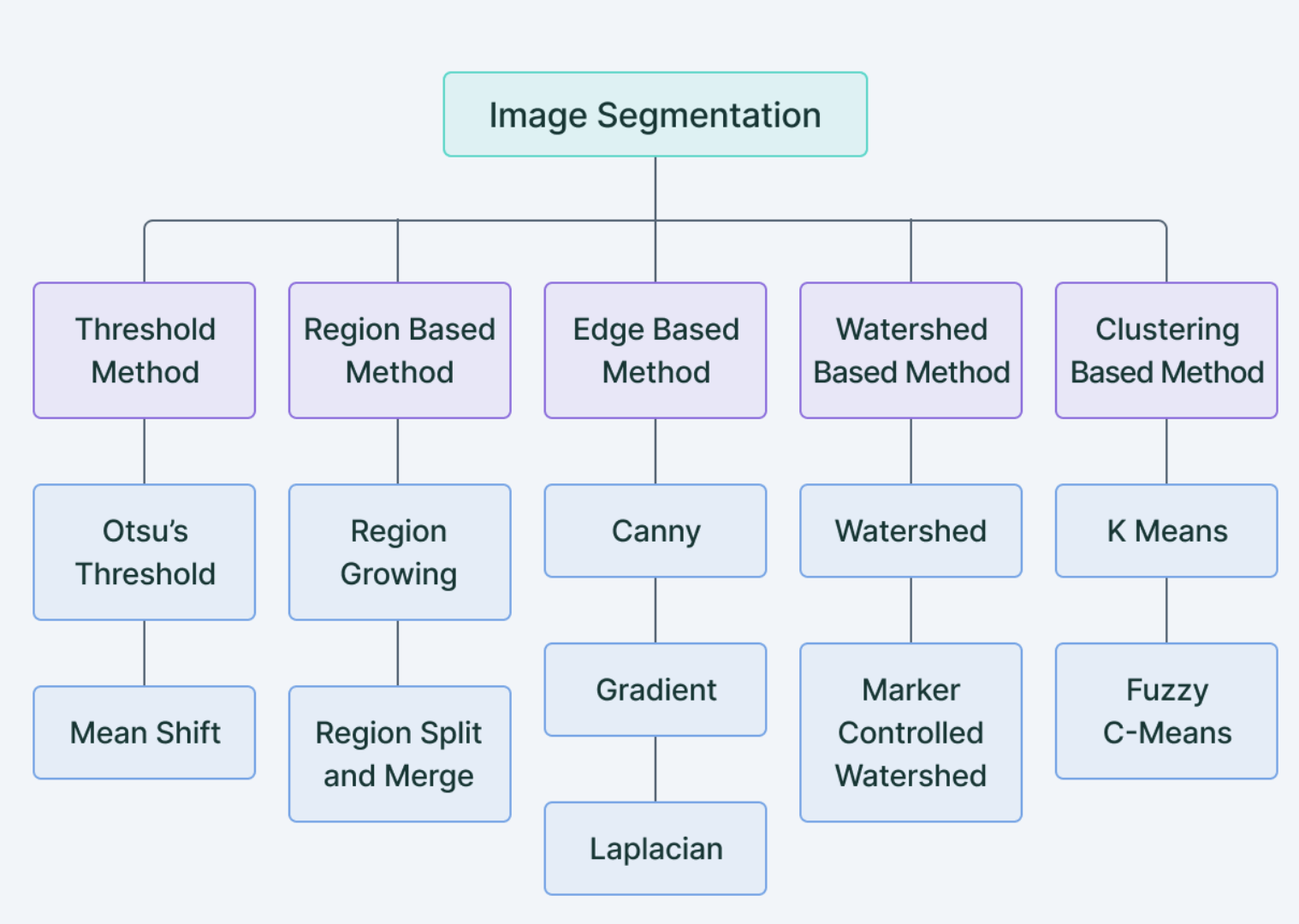}
    \caption{Types of image segmentation \cite{guide}}
    \label{fig:img_seg_types}
\end{figure}

In a study of \cite{ho}, these methods were categorized into four main types: edge-based, clustering-based, region-based, and split and merge approaches, each with specific algorithms. Another article classified methods into five subgroups, including threshold-based and watershed-based methods and their corresponding algorithms, as shown in Figure \ref{fig:img_seg_types}.


Before starting semantic segmentation of the plate, we must separate things from stuff. That means we detached the foreground plate from the background. It will optimize the number of pixels for \textit{Region-Growing algorithm}  that will serve on grouping clustered colors. 

We begin by analyzing an experimental image and grouping similar colors using a chosen color model. Using the Region-Growing algorithm, we define boundaries for these color clusters. To address edge clustering, we apply the Region-Merging algorithm (\textbf{Algorithm} \ref{alg}), merging smaller groups with nearby larger regions. This process yields semantically segmented objects.



\textit{ K-means clustering}

The k-means clustering algorithm partitions a dataset into distinct groups based on similarity to cluster centroids. It proceeds in the following steps:

\begin{itemize}
    \item \textit{Initialization.} Randomly select k initial cluster centroids from the dataset or use predefined initial values.
    \item \textit{Assignment} of each data point to the cluster whose centroid is closest. The distance is  found using the Euclidean distance formula:
    \[ d(x_i, c_j) = \sqrt{\sum_{p=0}^{n}(x_{i_p} - c_{j_p})^2} \]
    where \(d\) is the distance between data point \(x_i\) and centroid \(c_j\), \(n\) is the number of dimensions or features in the dataset, and \(x_{i_p}\) and \(c_{j_p}\)  are the \(p\)-th feature values of  \(x_i\)and  \(c_j\), respectively.
    \item \textit{Recalculate} the centroids of each cluster by taking the mean of the feature values of all data points assigned to that cluster.
    \item \textit{Repeat }steps 2 and 3 until convergence: Iterate steps 2 and 3 until the centroids no longer change significantly or until a predefined number of iterations is reached.
\end{itemize}

The k-means algorithm aims to minimize the within-cluster sum of squares, also known as the inertia, which measures the overall compactness of the clusters. The objective function to minimize can be expressed as:
\[J = \sum_{j=1}^{k}\sum_{x_i\subseteq C_j}^{}d(x_i, c_j)^2\]
where \(J\) is the total sum of squared distances, \(k\) is the number of clusters, \(C_j\) is the set of data points belonging to cluster \(j\), \(x_i\) is a data point, and \(c_j\) is the centroid of cluster \(j\).

The k-means algorithm converges when centroids stabilize, and data points remain assigned to clusters. Multiple runs with different initializations are often performed to find the best solution.

\vspace{12pt}
\textit{Getting the palette, then mask objects through them}

After color segmentation, we obtain a palette of colors matching the clusters. Using this palette, we generate masks by filtering the image with each color from the k-means algorithm.

\vspace{12pt}
\textit{Iterate each mask and determine it with classification}

With the color masks ready, we overlay each onto the original camera image. Since a color like red might appear in different objects, we analyze each object within the mask to assign its class.


Regions $R_k$ are considered to be sets of points with the following properties:
\begin{itemize}
    \item $x_i$ in a region R is \textit{connected} to $x_j$ if there is a sequence \{$x_i$, ..., $x_j$\} such that $x_k$ and $x_{k+1}$ are connected, and all the points are in R. (1)
    \item R is a \textit{connected region} if the set of points x in R has the property that every pair of points is connected. (2)
    \item I, the entire image = $\bigcup_{k=1}^{m} R_k$ (3)
    \item $R_i \bigcap R_j = \phi, i \neq j$ (4)
\end{itemize}
A set of regions satisfying (2) through (4) is known as a \textit{partition} \textcolor{OliveGreen}{\cite{ballard1982computer}}.

\subsubsection{Image Classification}
Food items are classified based on nutritional content using a classification algorithm, which categorizes data into predefined categories based on features. This involves learning from a labeled training dataset and predicting class labels for new data.

For an experiment of classification objects on the plate, we used the Support Vector Machine (SVM) algorithm.

\begin{algorithm}
 \KwData{Some initial region split image $M$ }
 \KwResult{Region Merged image}
 \SetKwProg{Fn}{Function}{}{}
 initialization\;
    $\rightarrow$ From initial regions in the image\;
    $\rightarrow$ Build a regions adjacency graph (RAG)\;
    \While{until no regions are merged}{
          Consider its adjacent region and test to see if they are similar\;
          For similar regions, merge them and modify the RAG\;
         }
    \Return $Merged \: colored \: image \: array$;
 \caption{Pseudocode of Region Merging algorithm}
\label{alg}
\end{algorithm}

\textit{Support Vector Machine classification algorithm}

SVM involves finding an optimal hyperplane that maximally separates different classes in the input data. Here is a step-by-step explanation:

\begin{enumerate}
    \item Data Representation. Like an image, each piece of input data represents a set of features. These features capture important details, like pixel values.
    \item Class Separation. SVM looks for a line or boundary that best separates different classes of data points. It tries to maximize the distance between this line and the closest data points of each class.
    \item Linear Separation. When a straight line can separate the data, SVM finds the best equation, called a hyperplane, like $w^Tx + b = 0$,
    where $w$ is the weight vector perpendicular to the hyperplane, and $b$ is the bias term. 
    \item Non-linear Separation. For more complex data that can't be separated with a straight line, SVM uses a "kernel trick" to transform the data into a higher-dimensional space where linear separation is possible.
    \item Margin Maximization. SVM selects the hyperplane that maximizes the space between classes. This helps improve accuracy and handles noisy data.
    \item Classification. Once the hyperplane is chosen, SVM can classify new data points by seeing which side of the line they fall on.
    \item Soft Margin. In cases where perfect separation isn't possible, SVM allows for some misclassifications with a penalty. This helps achieve better overall separation.
\end{enumerate}

By iteratively optimizing the hyperplane parameters, SVM can effectively classify data points into different classes while maximizing the margin of separation.

\subsection{Nutritional Assessment Algorithm}

In this subsection, we'll dive into how our system determines how healthy a meal is by looking at the balance of different food groups.

After obtaining the classes of the detected objects, we utilized the pixels of the image to assess their relevance. This involved highlighting each object's mask and calculating the percentage of that particular object in relation to the total food on the plate. By knowing the object type, we combined them and computed the overall quantity of fruits, vegetables, proteins, and whole grains:

\begin{equation}
    classType_i = \frac{\sum_{contour_i}^{n}(\sum_{j=0}^{m}(pixel))}{\sum_{k=0}^{total\_plate}(pixel)}
\end{equation}

We categorize data into four elements using a Python dictionary, as shown in Figure \ref{fig:categories}. Food is classified based on how closely it aligns with Harvard's recommendations—half the plate for fruits and vegetables and a quarter each for proteins and whole grains. The categories range from "Healthy food" (0-25\% error margin) to "Not a healthy plate" (exceeding 75\% error). Items not on our predefined list are considered junk food.

The balance level of a plate, denoted by \(B\), takes into account the contributions from fruits (\(f\)), vegetables (\(v\)), healthy proteins (\(hp\)), and whole grains (\(wg\)), with specified caps for each category:
\begin{equation}
    B= \min(f + v, 50) + \min(hp, 25) + \min(wg, 25)
\end{equation}
Here, \(B\) represents a plate's overall healthy balance level. The contributions from fruits and vegetables are combined and capped at 50, while the contributions from healthy proteins and whole grains are individually capped at 25.

To calculate the healthy products out of all products on the plate, we use the following formula:

\begin{equation}
    \text{H} = \frac{f + v + hp + wg}{T}
\end{equation}


In this formula, \(f\), \(v\), \(hp\), and \(wg\) represent the quantities of fruits, vegetables, healthy proteins, and whole grains on the plate, respectively. \(T\) is the total quantity (or percentage) of all products on the plate. This fraction represents the proportion of the plate that is made up of healthy products.

\section{Results}

\begin{figure}[tb]
    \centering
    \includegraphics[width=.4\textwidth]{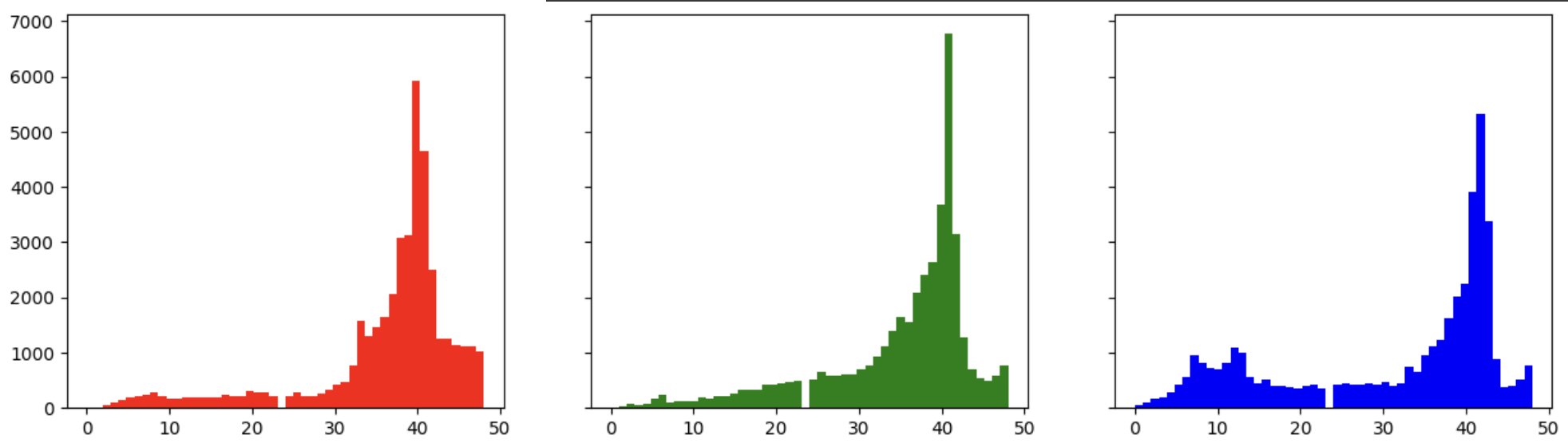}
    \caption{Color histogram of the initial plate image}
    \label{fig:color_hist}
\end{figure}

We tried various techniques like clustering, watershed analysis, and threshold-based methods during our experiments. Interestingly, the methods based on thresholds and clustering, specifically using the mean shift and K-means algorithms, showed the most promising results. We selected an image similar to that shown at the top of Figure \ref{fig:cie_lab} for segmentation. In Figure \ref{fig:color_hist}, you'll find a color histogram displaying the distribution of colors across different RGB channels in that image. In this section, we'll start by discussing classification metrics and then delve into the process of semantic segmentation. Finally, we'll present some user-friendly examples to illustrate how our approach can be applied in real-life scenarios.

\subsection{Accuracy Evaluation} 

\begin{figure}[bt]
    \centering
    \begin{subfigure}{0.15\textwidth}
        \centering
        \includegraphics[width=\textwidth]{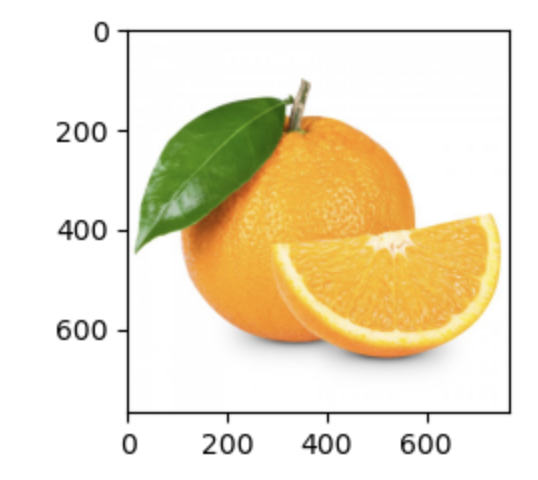}
        \caption{Original image for classification}
        \label{fig:orig_img_class}
    \end{subfigure}
    \hfill
    \begin{subfigure}{0.15\textwidth}
        \centering
        \includegraphics[width=\textwidth]{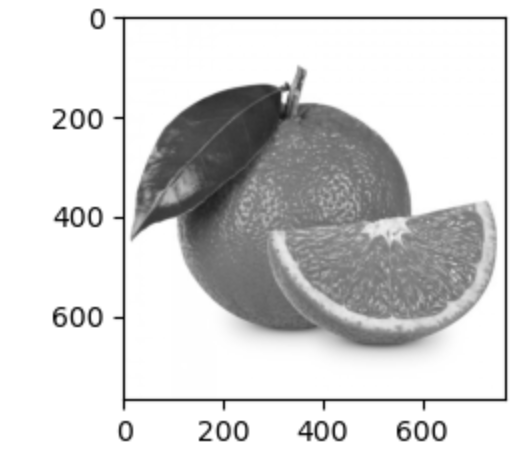}
        \caption{Grayscale color converted image}
        \label{fig:gray_img_class}
    \end{subfigure}
    \hfill
    \begin{subfigure}{0.15\textwidth}
        \centering
        \includegraphics[width=\textwidth]{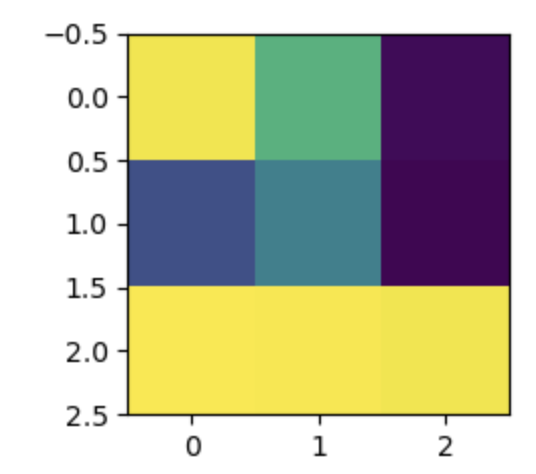}
        \caption{Centroids after K-Means algorithm}
        \label{fig:centroids}
    \end{subfigure}
    
    \medskip
    \begin{subfigure}{0.15\textwidth}
        \centering
        \includegraphics[width=\textwidth]{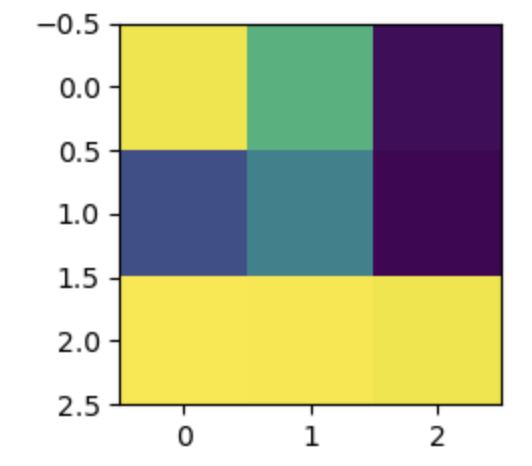}
        \caption{Quantization palette}
        \label{fig:quan_palette}
    \end{subfigure}
    \begin{subfigure}{0.15\textwidth}
        \centering
        \includegraphics[width=\textwidth]{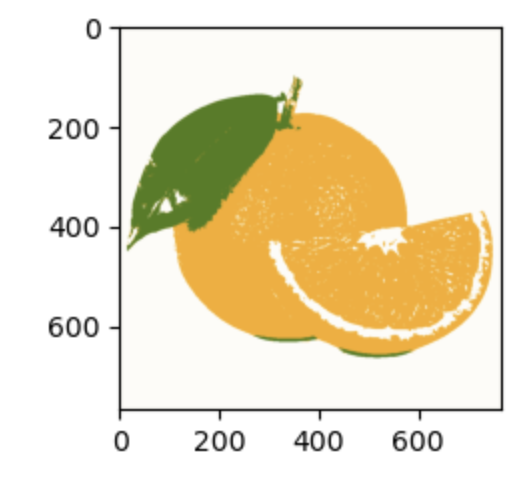}
        \caption{Quantified orange image}
        \label{fig:quantified}
    \end{subfigure}
    
    \caption{Stages of classification descriptor}
    \label{fig:classification_feature}
\end{figure}

\begin{figure}
    \centering
    \includegraphics[width=.45\textwidth]{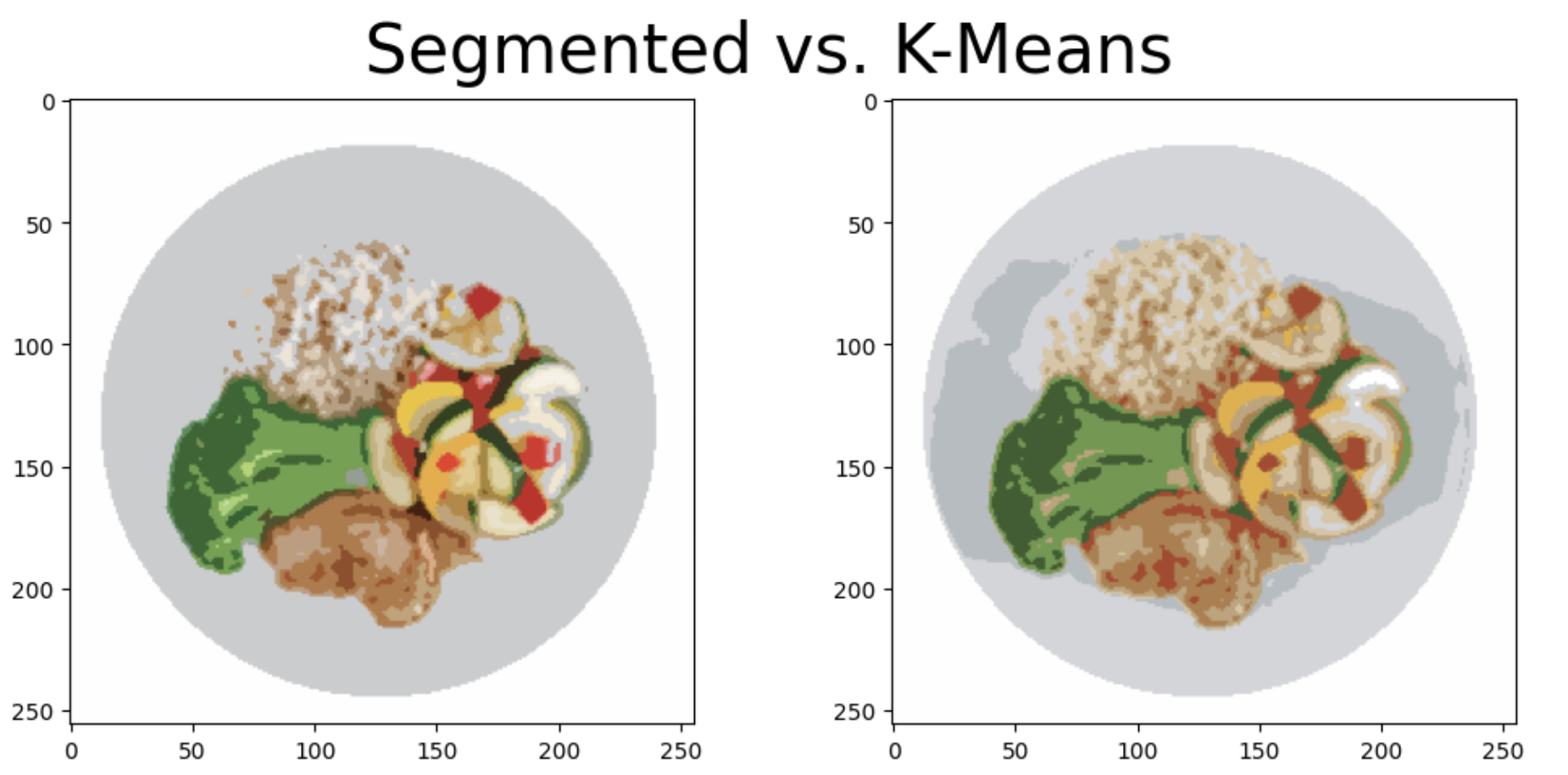}
    \caption{Region Based segmentation with Blurring vs. K-Means}
    \label{fig:from_blurring}
\end{figure}

We evaluated the performance of our system using two subsystems: classification and segmentation. The results of the current classification and segmentation operations are depicted in Figures \ref{fig:classification_feature}-\ref{fig:from_blurring}. For the segmentation algorithm, we utilized the resources from \cite{py-clustering} and \cite{py-feature-clustering}, while \cite{py-script} served as a valuable asset for the classification task. These algorithms proved to be efficient in implementing and extracting objects. We used Python, OpenCV \cite{opencv}, and scikit-learn \cite{scikit}  libraries. 


A primary metric of classification as precision scores can be observed in Table \ref{table:svm_result}, with an accuracy of 0.83 for the apple and 0.86 for the orange.

\begin{table}[tb]
\centering
\begin{tabular}{ |c|c|c|c| } 
\hline
       & Precision & Recall & Accuracy \\
Apple  & 0.833     & 1      & 0.833    \\
Orange & 1         & 0.857  & 0.857   \\
\hline
\end{tabular}
\caption{Measurements of SVM classifier on apple and orange.}
\label{table:svm_result}
\end{table}



Fig. \ref{fig:classification_feature} illustrates the individual stages of the manifestation of features in the image. The centroid and palette are important indicators for object recognition, obtained by applying the K-Means algorithm built into OpenCV. In addition, the aspect of color conversion is of great importance since the use of the HSV color space has revealed additional distinctive characteristics. However, the segmentation algorithm showed sub-optimal performance, which led to a decrease in the achieved results.

\begin{figure}[bt]
    \centering
    \includegraphics[width=.35\textwidth]{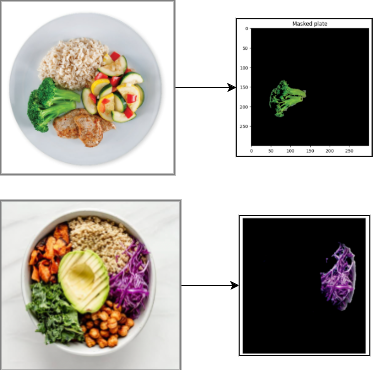}
    \caption{Mask examples. Broccoli and red cabbage extraction using Region-Based algorithm}
    \label{fig:cie_lab}
\end{figure}

We ran an experiment using the sklearn K-Means algorithm to group similar parts of an image together; then, we tried to isolate specific areas using region-expanding. After a few tries, we isolated broccoli from a meal, as shown in Figure \ref{fig:cie_lab}. When we used Mean-Shifting segmentation, the resulting clusters weren't very connected compared to K-Means, which worked better overall, as seen in Figure \ref{fig:cie_lab}.
We also found that the CIE LAB color space didn't work well for certain foods, like rice, because it treated shadows and contrasts as separate colors. On the other hand, HSI color space did a better job keeping the rice together, as you can see in the left part of Figure \ref{fig:cie_lab}. To achieve this, we first used K-Means to group similar colors, then expanded those groups into areas and created a mask. With this mask, we filled in the gaps in the original image and determined the dominant color for each masked area.

The left image of Fig.\ref{fig:from_blurring} shows the final image segmentation stage. As described in the \ref{sec:methodology}, after the K-Means algorithm, we determined the appropriate color palette. Subsequently, this mask was used to identify the segmented object in the original image. 


\subsection{Prototype Application} 
 The Healthy Plate evaluation method can be used in various useful applications, such as diet or restaurant recommendation systems for individuals seeking healthy options. Identifying nutritious options from a menu filled with a great variety of choices can be a daunting task. Imagine a world where restaurant recommendations consider taste and prioritize your well-being. Our algorithm enables us to evaluate the healthiness of restaurant options, transcending traditional recommendations' boundaries. By analyzing menus and identifying the healthiest options, we empower individuals to make choices aligned with their dietary goals.


 
\begin{figure}[bt]
    \centering
    \includegraphics[width=.4\textwidth]{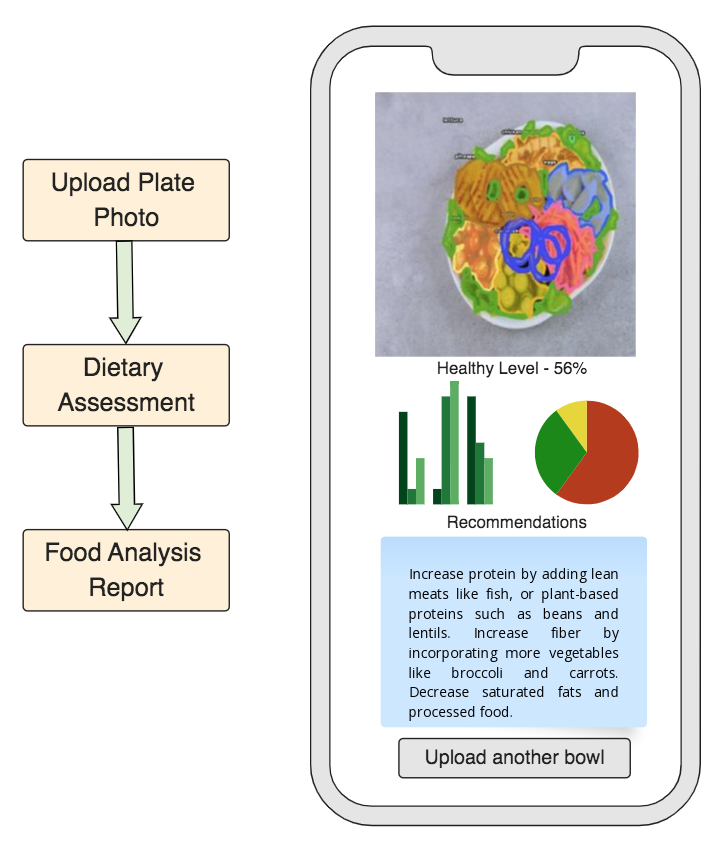}
    \caption{Prototype of the system }
    \label{fig:app}
\end{figure}

\begin{figure}[bt]
    \centering
    \includegraphics[width=.4\textwidth]{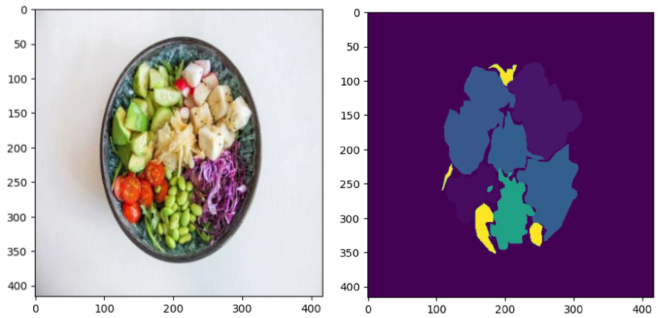}
    \caption{Evaluation example. Original and masked images. Calculated Balance level $B$ is 51, Healthy level $H$ is 99.}
    \label{fig:evex}
\end{figure}

The mockup of the prototype application is presented in Fig. \ref{fig:app}. Users can interact with the system through a mobile app or web interface to get dietary assessments and restaurant recommendations. The evaluation example is presented in Fig \ref{fig:evex}. Calculated Balance level $B$ is 51, Healthy level $H$ is 99.

\section{Conclusion}
The problems of inadequate eating in modern society can be addressed by Harvard's research on the Healthy Eating Plate. By emphasizing the consumption of whole, unprocessed foods and balanced nutrient intake, the Healthy Eating Plate offers evidence-based guidelines to address inadequate eating habits.

The presented method pioneers the application of image processing techniques to dissect plates into components—vegetables, proteins, and carbohydrates—offering. As a result, a plate's nutritional profile is based on recent Harvard studies. Besides empowering people to make healthier choices, our system contributes to a broader societal shift towards conscious eating. Such applications potentially impact dietary habits and health and can serve as a brick in integrating AI with nutritional science to promote healthier lifestyles.

The proposed system has certain limitations. One of the key issues we encountered was the inaccurate segmentation of objects, leading to difficulties in accurately recognizing and identifying overlapping objects. Moreover, there are still challenges related to food classification due to various cooking methods and sauces. As for further research or improvements to the system, we plan to enhance the segmentation process to achieve more precise object separation. 

\Urlmuskip=0mu plus 1mu\relax
\bibliographystyle{vancouver}
\bibliography{library}

\end{document}